\RequirePackage{snapshot}
\documentclass[runningheads]{llncs}
\usepackage{url}
\usepackage[T1]{fontenc}
\usepackage{graphicx}

\usepackage[acronym]{glossaries}
\usepackage{glossaries-extra}
\setabbreviationstyle[acronym]{long-short}
\newacronym{tsp}{TSP}{Traveling Salesperson Problem}
\newacronym{ai}{AI}{Artificial Intelligence}
\newacronym{ga}{GA}{Genetic Algorithm}
\newacronym{cx}{CSX}{Circular Shift Crossover}
\newacronym{rx}{RX}{Reversal Crossover}
\newacronym{crx}{CSRX}{Circular Shift Reversal Crossover}
\newacronym{box}{BOX}{Best Order Crossover}
\newacronym{ox}{OX}{Order Crossover}
\newacronym{opt}{OPT}{Optimal Tour Length}

\usepackage{hyperref}
\usepackage{csquotes}
\usepackage{cleveref}

\usepackage{caption}
\usepackage{subcaption}

\usepackage{booktabs}

\usepackage{tikz}
\usepackage{color}
\usepackage{amsmath,amsfonts,amssymb}

\newcommand{\uram}[1]{\marginpar{\tiny\textcolor{blue}{Martin:  #1}}}
\newcommand{\sthu}[1]{\marginpar{\tiny\textcolor{violet}{StHu:  #1}}}
\usepackage[left]{lineno}
\renewcommand{\uram}[1]{}
\renewcommand{\sthu}[1]{}

\begin{document}
	\title{CSRX: A novel Crossover Operator for a Genetic Algorithm applied
			to the Traveling Salesperson
			Problem\thanks{This preprint has not undergone peer review or any
			post-submission improvements or corrections.
			The Version of Record of this contribution is published in
			Haber, P., Lampoltshammer, T.J., Mayr, M. (eds) Data
			Science--Analytics and Applications. iDSC 2023, and is
			available
			online at
			\url{https://doi.org/10.1007/978-3-031-42171-6_3}}
	}
	\titlerunning{CSRX: A novel Crossover Operator for a GA applied
					to the TSP}

	\newboolean{anonymized}

	\setboolean{anonymized}{false}

	\ifthenelse{\boolean{anonymized}}
	{
		\author{First Author
		\and
		Second Author
		\and
		Third Author
		}
		\authorrunning{F. Author et al.}
		\institute{
			Institute, University, City, Country\\
			\email{\{first, second, third\}@uni.com}
			}

	}{
	\author{Martin Uray\inst{1,2}%
			\and
			Stefan Wintersteller\inst{2}
			\and
			Stefan Huber\inst{1,2}%
	}
	\authorrunning{M. Uray et al.}
	\institute{Josef Ressel Centre for Intelligent and Secure Industrial
	Automation,\\
	Salzburg University of Applied Sciences, Salzburg, Austria\\
		\email{\{
			martin.uray,
			stefan.huber
			\}@fh-salzburg.ac.at}
	\and
	Department for Information Technologies and Digitalisation,\\
		Salzburg University of Applied Sciences, Salzburg, Austria\\
	\email{ swintersteller.its-m2020@fh-salzburg.ac.at}}
	}
	\maketitle              %
	\begin{abstract}

		In this paper, we revisit the application of \gls{ga} to the
		\gls{tsp} and introduce a family of novel crossover operators that
		outperform the previous state of the art.
		The novel crossover operators aim to exploit symmetries in the
		solution space, which allows us to more effectively preserve
		well-performing individuals, namely the fitness invariance
		to circular shifts and reversals of solutions.
		These symmetries are general and not limited to or
		tailored to \gls{tsp} specifically.

	\end{abstract}

	\section{Introduction}\label{sec:introduction}

  Given $n$ points in the plane, the \glsfirst{tsp} asks for the shortest closed
  tour that visits all points. It is a classical optimization problem and known
  to be NP-complete. Consequently, besides approximately optimal algorithms of
  polynomial complexity (PTAS), it is also a natural candidate for all kinds of
  optimization methods in \gls{ai}. In particular, essentially the entire
  palette of meta-heuristic methods has been thoroughly studied for \gls{tsp},
  including \glsfirst{ga}.

  \gls{ga} is used for optimization and search tasks and inspired by natural
  selection in evolution.
  Given a set of candidate solutions, fit individuals are selected and
  recombined to form fitter offsprings, such that a population of individuals
  evolves over time from generation to generation. The recombination happens by
  means of a so-called crossover operator.

	In 2021, the first competition on solving the \gls{tsp} was organized.
	This competition focused on the application of surrogate-based
	optimization and Deep Reinforcement Learning.
	For an overview of the
	literature on both domains, including an overview of the application of
	Deep Neural Networks, as well as a description of the competing
	methods, the reader may be referred to~\cite{BdCA+22}.

	For special variants of the \gls{tsp}, adaptions of the \gls{ga}
	were proposed to solve, among others, large-scale colored
	\gls{tsp}~\cite{Don2019}, or real-life oriented multiple
	\gls{tsp}~\cite{Lo2018}.
	Crossover operator tailored for the \gls{tsp} are introduced for
	multiparent~\cite{Roy2019} and
	multi-offspring~\cite{Wang2016} setups.
	Based on the cycle crossover, a modified cycle crossover operator is
	introduced for the application with the \gls{tsp}~\cite{Hus2017}.
	A more comprehensive overview of the \gls{ga} applied to the \gls{tsp} is
	given by Osaba et al.~\cite{OYD20}.
	There, the most notable crossover variants for the \gls{tsp} are being
	discussed.

	Gog and Chira~\cite{Gog2011} introduced the \gls{box}, a crossover
	operator extending the well-established \gls{ox}~\cite{Dav1991}.
	The \gls{ox}s intended application are order-based problems.
	For crossing over two parents, the \gls{ox} chooses two random splitting
	points within the first parent and copies the enclosed strings into the
	first offspring.
	The remaining numbers are then inserted in the order they appear in the
	second parent.
	On the other hand, the \gls{box} operator incorporates the knowledge about
	the global best individual to generate a new offspring.
	By using random splitting points and assigning each a value, the
	offspring is generated by using the numbers from the first parent,
	defined by the order of the segments of the second parent and the global
	best.
	In the comparative study~\cite{Gog2011}, the \gls{box} outperforms all
	other operators.

  We observe that \gls{tsp} possesses certain symmetry properties. However,
  well-known crossover operators do not respect these and, as a consequence,
  fail to produce fit offsprings from fit parents. Following this observation,
  we present a novel family of crossover operators that are designed to respect
  these symmetries and demonstrate that our novel crossover operators
  significantly outperform the previous state of the art.%
  Since the symmetries we exploit are of general nature, our methodology is of
  general interest and not limited to \gls{tsp} specifically.

  \section{Background}\label{sec:background}

  \subsection{\glsfirst{tsp}}

  Informally, \gls{tsp} asks for the shortest, closed path that visits a given
  set of $n$ locations exactly once. Note that we can reduce \gls{tsp} to
  find the optimal order of the locations $p_0, \dots, p_{n-1}$ that yields
  the shortest length of the resulting tour.

  To encompass a large variety of applications, we can formalize \gls{tsp} as
  follows: Given $p_0, \dots, p_{n-1}$ in a metric space $(X, d)$ with a metric
  $d$, encode a tour as a permutation $\pi \colon \{0, \dots, n-1\} \to \{0,
  \dots, n-1\}$ and ask for a tour $\pi$ that minimizes the tour length
  $\ell(\pi)$ with
	\begin{align}
    \ell(\pi) = \sum_{i = 0}^{n-1} d(p_{\pi(i)}, p_{\pi((i+1) \bmod n)}).
	\end{align}
	In this paper, we may interchangeably represent a permutation $\pi$ as
	the sequence $(\pi(0), \dots, \pi(n-1))$, when it fits better to the
	formal setting.

  A natural choice for $d$ is the Euclidean metric, which we use for
  experiments in this paper. However, the discussed methods work with any
  metric. For instance, to optimize the commissioning sequence in high-bay
  storage, the movements in the vertical and horizontal direction may not happen
  simultaneously or at the same speed, which can be modeled by $d$ accordingly,
  e.g., using the Manhattan metric or an affine transform of the Euclidean
  metric.

  In literature, there is a special asymmetric variant of the
  \gls{tsp}~\cite{AGMG+17}.
  This variant is characterized by the property, that the distances have a
  different value, dependent on their traversal direction, i.e. where a tour
  $d(p_{\pi(i)}, p_{\pi(i+1)}) \neq d(p_{\pi(i+1)}, p_{\pi(i)})$.
  Nevertheless, in this work, we only consider the vanilla, symmetric \gls{tsp}.

	\subsection{\glsfirst{ga}}\label{ssec:ga}

  The \gls{ga} is an optimization and search procedure that is inspired by the
  maxim \enquote{survival of the fittest} in natural evolution. A candidate
  solution (individual) is encoded by a string (genetic code) over some
  alphabet. Individuals are modified by two genetic operators: (i) random
  alteration (mutation) of single individuals and (ii) recombination
  (crossover) of two (or more) parents to form offsprings.
	Given a set of individuals
	(population), a selection mechanism based on a fitness function
	together with the two genetic operators produce a sequence of
	populations (generations).
	The genetic operators promote exploration of
	the search space while the selection mechanism attempts to promote the
	survival of fit individuals over generations.

	For \gls{ga} to work well, it is paramount that a suitable genetic
	representation of individuals is used.
  In particular, the crossover operator needs to have the property that the
  recombination of the genetic codes of two fit parents produces fit offsprings
  again, otherwise the genetic structure of fit individuals would not survive
  over generations and \gls{ga} easily degenerates to a randomized search. (We
  shall also note that the crossover operator also needs to produce valid
  genetic encodings of individuals, e.g., encodings of permutations for
  \gls{tsp}.)
  For more details on the \gls{ga}, the reader is referred
  to~\cite[Chapter 4]{RN2021}.

  For \gls{tsp}, it is common to take the sequence $(\pi(0), \dots, \pi(n-1))$
  of a permutation $\pi$ as the genetic code and $-\ell(\pi)$ would be the
  fitness of $\pi$.
  The so-called one-point crossover operator works as follows for \gls{tsp}. It
  takes two parents $\pi_1$ and $\pi_2$, chooses a random split index $0 \le s
  < n$, and produces an offspring $\pi_1 \times_s \pi_2$ as follows: It takes the
  prefix $(\pi_1(0), \dots, \pi_1(s))$ of $\pi_1$ and fills up the remaining
  entries in the order as they occur in $\pi_2$, to form a permutation $\pi_1
  \times_s \pi_2$. This way the \enquote{genetic information} of both parents
  is recombined.
  For instance, let $\pi_1 = (0,\ldots, 7)$ and $\pi_2 = (7,\ldots, 0)$ and let
  $s = 3$. Then $\pi_1 \times_3 \pi_2 = (0, 1, 2, 3, 7, 6, 5, 4)$. Besides this
  elementary crossover operator, there are many more. We refer
  to~\cite{Gog2011} for comparison for \gls{tsp}.

	\section{Novel crossover operators exploiting symmetry}\label{sec:contrib}

  Let us again consider the example of $\pi_1 \times_s \pi_2$ and let us assume
  that $\pi_1$ was a fit individual, or for the sake of argument, let us assume
  that $\pi_1$ is indeed optimal. Observe that $\pi_2$ is the reversed
  version of $\pi_1$, which we denote by $\pi^*_1$, and hence has the same
  fitness. Though $\pi_1$ might be optimal, $\pi_1 \times_s \pi^*_1$ is in
  general far from optimal, since we first traverse $\pi_1$ and after index $s$
  we reverse direction due to $\pi^*_1$. A similar situation happens when
  $\pi_2$ is a circularly shifted version of $\pi_1$.
  We observe that the one-point crossover operator is ignorant of the
  symmetries of \gls{tsp}, and so are all crossover operators reviewed
  in~\cite{Gog2011}. This, however, impairs the
  preservation of fit solutions or substructures over generations.

  From a more algebraic point of view, let us denote by $\Pi$ the set of all
  permutations of $\{0, \dots, n-1\}$. Note that \gls{tsp} possesses the
  following symmetry properties: A permutation $\pi \in \Pi$ or its reversed
  counterpart, or circular shifts of either, are essentially the same
  solutions, not the least according to the fitness. Let us denote by $\equiv$
  the corresponding equivalence relation over $\Pi$, i.e., $\pi \equiv \pi'$ if
  they are the same modulo reversing and circular shifting. So instead of
  \gls{ga} to act on the original space $\Pi$, we want to act on the smaller
  quotient space $\Pi / \equiv$, which more concisely captures the problem
  \gls{tsp}.

  More precisely, we want a new crossover operator $\overline{\times}$ that
  acts on $\Pi / \equiv$ in the following sense: If $\pi_1 \equiv \pi'_1$ and
  $\pi_2 \equiv \pi'_2$ then we also want $\pi_1 \overline{\times} \pi_2 \equiv
  \pi'_1 \overline{\times} \pi'_2$. That is, $\overline{\times}$ agrees with
  $\equiv$ and we can think of $\overline{\times}$ acting on
  congruence classes of $\equiv$, e.g., $[\pi_1]_\equiv \overline{\times}
  [\pi_2]_\equiv$ is justified as notation.

  We achieve this by, in some sense, factoring out circular shifts and
  reversals of individuals during the crossover.
  When we factor out the circular shifts for the one-point crossover operator,
  we call this the \gls{cx} operator. When we factor circular shifts and
  reversals, we call this the \gls{crx} operator. The \gls{rx} operator, we
  introduced in a preliminary work in~\cite{WULH22}, can be interpreted as
  factoring out reversals from the one-point crossover operator.
  However, in general, this technique can be applied to all kinds of existing
  crossover operators, including \gls{box}, which is the current
  state of the art~\cite{Gog2011}.

  \begin{figure}[b]
    \centering
    \includegraphics[width=4.5cm]{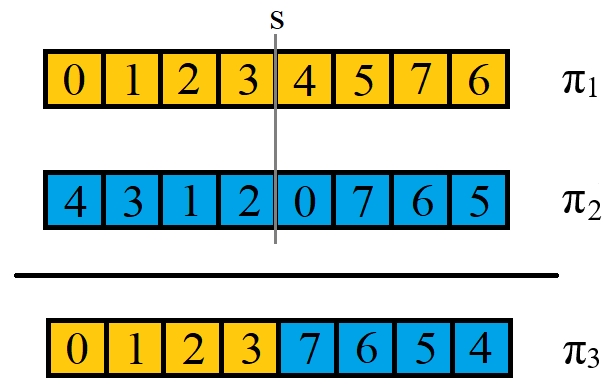}
    \caption{The recombination of $\pi_1$ and $\pi_2$ into an offspring
    $\pi_3$ using \gls{cx}.}
    \label{fig:csx}
  \end{figure}

  In more detail, \gls{cx} works as follows: Let us consider $\pi_1$ and
  $\pi_2$ and choose a split index $s$. Now circularly shift $\pi_2$ such that
  $\pi_2(s) = \pi_1(s)$ and apply the ordinary one-point crossover. See
  \cref{fig:csx} for an example.

\begin{figure}[t]
        \centering
        \begin{subfigure}[b]{0.4\textwidth}
            \centering
            \includegraphics[width=4.5cm]{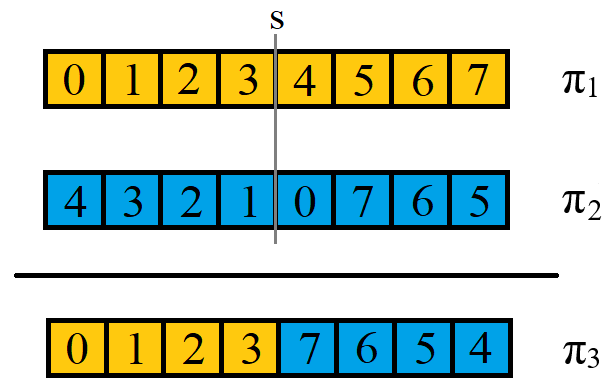}
            \caption{$\pi_1$ and $\pi_2$ recombined to candidate state
                $\pi_3$ using \gls{cx}.}
            \label{fig:crx1}
        \end{subfigure}
        \hfill
        \begin{subfigure}[b]{0.4\textwidth}
            \centering
            \includegraphics[width=4.5cm]{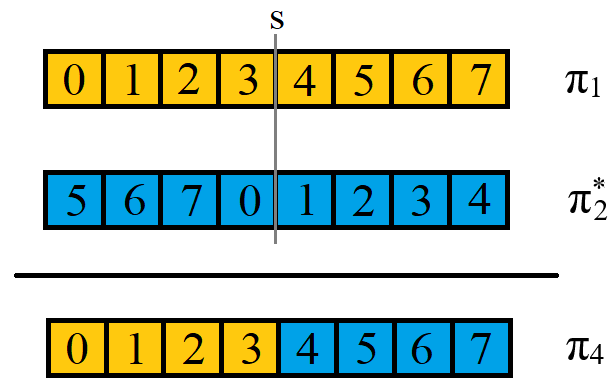}
            \caption{$\pi_1$ and $\pi^*_2$ recombined to candidate state
                $\pi_4$ using \gls{cx}.}
            \label{fig:crx2}
        \end{subfigure}

        \caption{Recombination of $\pi_1$ and $\pi_2$ to candidates $\pi_3$ and
          $\pi_4$ for \gls{crx}.
      }
        \label{fig:crx}
    \end{figure}

  For \gls{rx} we face the obstacle that we cannot \enquote{normalize} the
  traversal direction of a tour $\pi$ in natural way. That is, we cannot tell
  directly whether $\pi_2$ or $\pi^*_2$ is \enquote{compatible} with $\pi_1$,
  so we let the fitness decide: We consider both candidates of the one-point
  crossover, $\pi_1 \times_s \pi_2$ and $\pi_1 \times_s \pi^*_2$, and take the
  one with the better fitness. This way we get invariant of the traversal
  direction of $\pi_2$, and this is what we aim for from the algebraic point of
  view from above. For \gls{crx} we combine both strategies, i.e., we apply
  \gls{cx} to combine $\pi_1$ with $\pi_2$ and $\pi^*_2$ and take the one with
  the better fitness. See \cref{fig:crx} for an example. Note that $\pi_2$ is a
  reversed and circularly shifted version of $\pi_1$ in \cref{fig:crx} and
  $\pi_1$ is restored in \cref{fig:crx2}. That is, if $\pi_1$ was optimal then
  \gls{crx} again yields an optimal individual.

	\section{Experimental evaluation}%

  \paragraph{Setup.}
	For all the experiments in this work, a common setup is established.
	As there are already libraries for standard implementations of the
	\gls{ga} algorithm, we do not implement the algorithms
	from scratch, rather we use a library called
	\textit{mlrose}\footnote{\url{https://mlrose.readthedocs.io/}} as a
	basis and improve the above stated algorithm based on this library.
	This library already provides a mapping of the \gls{tsp} to a set of
	implementations of well-known \gls{ai} methods, which makes it a
	favorable candidate for our experiments.

    The proposed \gls{crx} is challenged against \gls{box}, which was the best
    performing crossover operator on \gls{tsp} as reported in~\cite{Gog2011}
    and we follow the implementation details given in~\cite{Gog2011}.

  The experiments are evaluated on three different
  standard data sets for \gls{tsp}: \textit{att48}, \textit{eil51} and
  \textit{st70}. All three are included in the TSPLIB~\cite{Rei91}. The data
  set \textit{att48} contains 48 cities in a coordinate system with a known
  \gls{opt} of $33523$. The \textit{eil51} data set has 51 cities with an
  \gls{opt} of $426$, and the \textit{st70} data set has 70 cities and an
  \gls{opt} of $675$.

  \paragraph{Results.}
	All experiments are configured using a shared parameter set, with a
	population size of $100$ and a mutation rate of $0.05$ as defined in the
	reference paper~\cite{Gog2011}.
  	An inversion-mutation is implemented as a mutation strategy~\cite{Fogel1993}.

	The first experiment uses a maximum number of $1000$ generations.
	To ensure more reliable results, the experiments were repeated $10$ times
	with different seeds.

	The original paper does not specify how exactly the elitism is configured.
	However, our experiments show that an elitism size of $10\%$ performs the best
	among all tested parameter configurations ($[0\%, 1\%, 10\%, 20\%, 30\%]$),
	and showed the highest convergence to the numbers presented
	in~\cite{Gog2011}.

	The results of this first experiment can be seen in \cref{tab:results1000}.
	This table illustrated the numbers (mean $\pm$ std) from
	the original paper
	(\gls{box}(orig)), the numbers from our reimplementation
	(\gls{box}(reimp)), and the introduced \gls{crx} operator,
	accompanied by the optimal tour length (OPT).
	Since the referenced paper~\cite{Gog2011} did not evaluate the
	\gls{box} operator towards the \textit{att48} data set, no numbers are
	available.
	We denote the relative error of \gls{box} by
	$\Delta_{rel}^{BOX} = \frac{BOX-OPT}{OPT}$, and
	$\Delta_{rel}^{CSRX}$
	analogously for \gls{crx}.

	\begin{table}[t]
		\centering
    \caption{Experimental results after $1000$
    generations.}\label{tab:results1000}
		\begin{tabular}{cllllll}
      \toprule
			data set & \gls{opt} & \gls{box}(orig) &
			\gls{box}(reimp) &
			\gls{crx} & $\Delta_{rel}^{BOX}$ & $\Delta_{rel}^{CSRX}$
			\\
      \midrule
			\textit{att48}               & $33523$   & n/a       & $35872\, (\pm 806)$ &
			\textbf{$34789\, (\pm 488)$} & $7.00\%$  & $3.77\%$ \\
			\textit{eil51}               & $426$     & $460$     & $460\, (\pm 9)$     &
			\textbf{$442\, (\pm 5)$}     & $7.98\%$  & $3.75\%$ \\
			\textit{st70}                & $675$     & $741$     & $818\, (\pm 24)$    & \textbf{$708\, (\pm
			15)$}                        & $21.19\%$ & $4.88\%$ \\
      \bottomrule
		\end{tabular}
	\end{table}

\begin{figure}[t]
        \centering
        \includegraphics[width=0.9\textwidth]{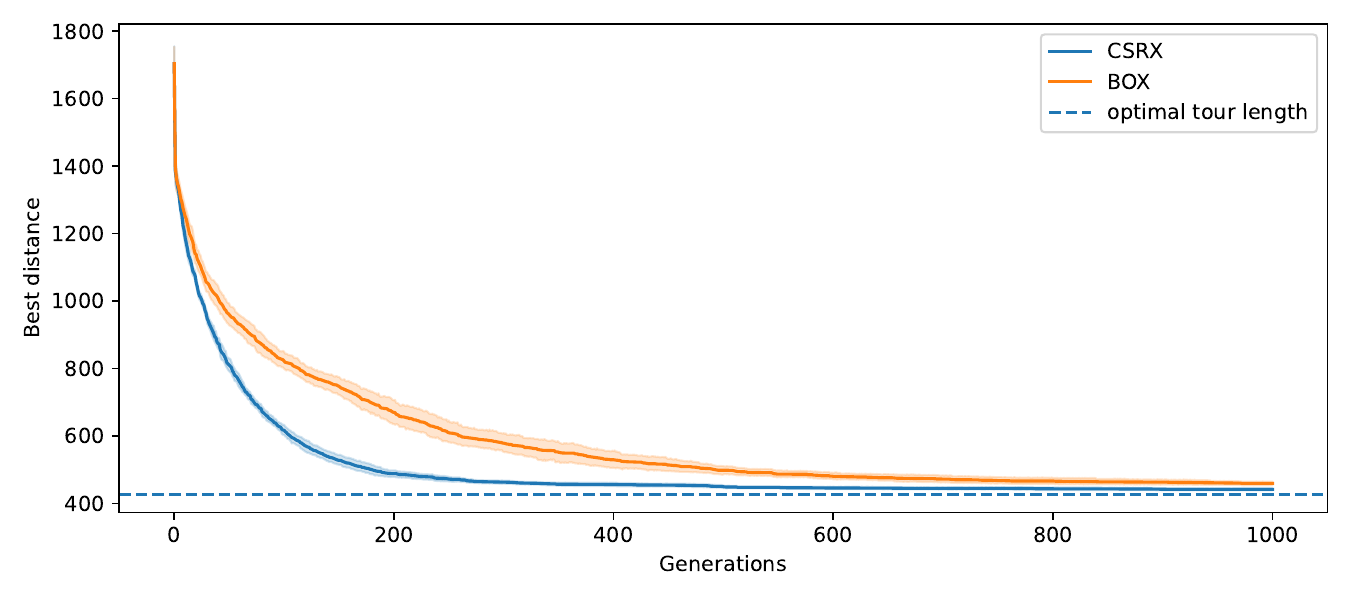}
        \caption{The results on a $95\%$ confidence
		interval. \gls{box}(reimp) compared with
		\gls{crx} on $1000$ Generations on the \textit{eil51} data set.
		This experiment replicates the results from~\cite{Gog2011}.}
        \label{fig:results1000}
    \end{figure}

	As can be seen in \cref{tab:results1000}, the reported numbers of
	\gls{box}(orig) could not be reproduced exactly for the
	\textit{st70} data set.\footnote{
		This issue was brought to the attention of the authors of the
		reference. We contacted the authors, but no response was received
		at the point of submission.}
	Even if this single number diverges, it can be seen that the introduced
	\gls{crx} operator outperforms both, \gls{box}(orig) and \gls{box}(reimp).
	On the \textit{att48}, \textit{eil51}, and \textit{st70} data sets,
	difference of the
	relative error ($\Delta_{rel}^{BOX} - \Delta_{rel}^{CSRX}$)
	of $3.23\%$, $4.23\%$, and $16.29\%$ are observed, respectively.
	On all three data sets, the \gls{crx} operator has a much lower standard
	deviation.
    \Cref{fig:results1000} shows the results over the course of the $1000$
	generations.

    As can be clearly seen in \cref{fig:results1000}, the \gls{crx} operator
    is outperforming the \gls{box} operator already within the first $200$
    generations.
    Based on this fact, the results for the following experiment are
    bootstrapped: we reduce the computational costs of our experiments by decreasing the
    number of generations to $200$ in the following.
    In turn, the number of experiments conducted has been raised to $100$.

\begin{figure}
        \centering
		\includegraphics[width=0.9\textwidth]{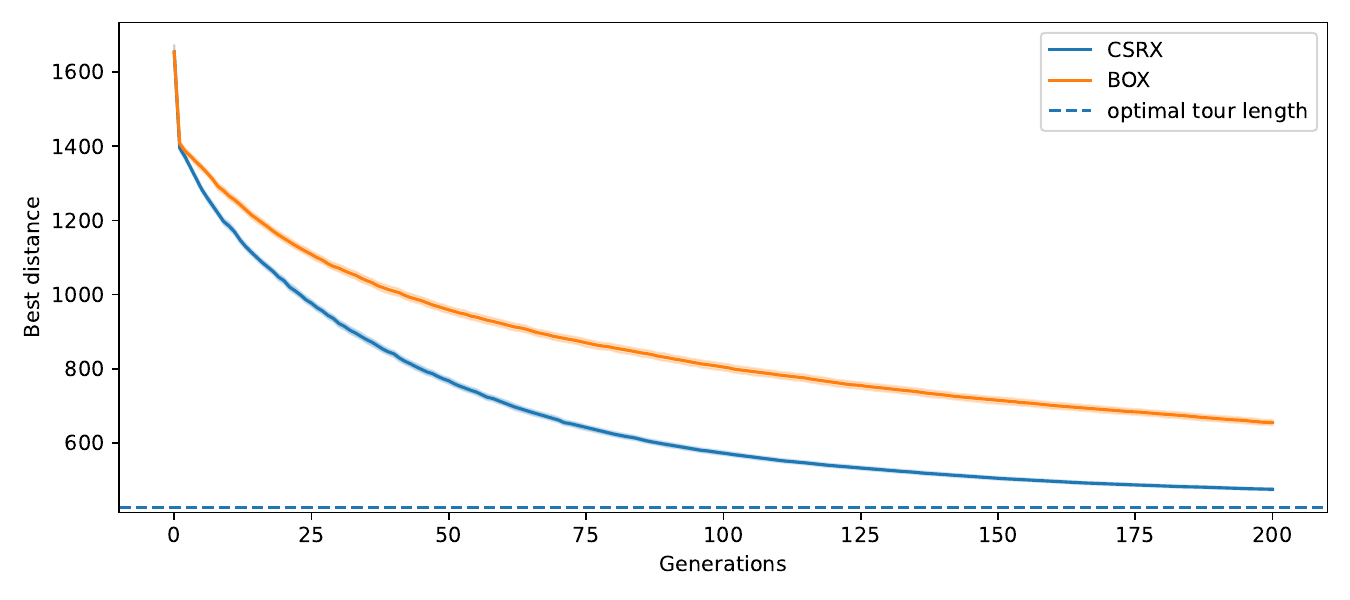}
		\caption{The second experiment on a $95\%$ confidence
		interval. The \gls{box}(reimp) compared with
		\gls{crx}
		on $200$ Generations on the \textit{eil51} data set.}
		\label{fig:results200}
	\end{figure}

	For this setup, an elitism of $20\%$ empirically performs best on
    both, the \gls{crx} and the \gls{box}, among the defined search space.
	The results of this experiment can be seen in \cref{tab:results200} and
    \cref{fig:results200}, numerically and visually over the number of
    generations, respectively.
	Also, these results show, that the introduced
	\gls{crx} operator outperforms the \gls{box} operator.
	The difference of the relative error on the three data sets are of $39
	.38\%$, $42.25\%$, and	$83.40\%$
	for the data sets \textit{att48}, \textit{eil51}, and \textit{st70},
	respectively.
	Also here, the standard deviation is significantly lower, compared to the
	\gls{box}.

	\begin{table}[t]
		\centering
    \caption{Experimental results after $200$
    generations.}\label{tab:results200}
		\begin{tabular}{clllll}
      \toprule
			data set & \gls{opt} & \gls{box}(reimp) & \gls{crx} &
			$\Delta_{rel}^{BOX}$ & $\Delta_{rel}^{CSRX}$
			\\
			\midrule
			\textit{att48} & $33523$    & $50033\, (\pm 3648)$  & $36830\, (\pm
			1233)$         & $49.25\%$  & $9.86\%$             \\
			\textit{eil51} & $426$      & $655\, (\pm 39)$      & $475\, (\pm
			15)$           & $53.76\%$  & $11.50\%$            \\
			\textit{st70}  & $675$      & $1484\, (\pm 85)$     & $921\, (\pm
			46)$           & $119.85\%$ & $36.44\%$            \\
      \bottomrule
		\end{tabular}
	\end{table}

	\section{Conclusion}
	\label{sec:conclusion-and-final-remarks}

  In this paper, we introduced a new technique of general nature on how to
  facilitate crossover operators, aiming to improve \gls{ga} on the task of the
  \gls{tsp}.
	These technique utilize the symmetry properties of the optimization problem
	so that two fit parents will not be steered to generate unfit offsprings.

  Furthermore, we used these proposed techniques to facilitate a very simple
  one-point crossover to demonstrate their abilities and received the
  \gls{crx} operator. In an experimental evaluation against the best performing
  crossover operator for the \gls{tsp} in the latest comparative
  study~\cite{Gog2011}, the \gls{box}, the \gls{crx} clearly outperformed on
  all given tasks. In future work we plan to transfer our technique to
  \gls{box}.

  \bibliographystyle{splncs04}
  \bibliography{tsp_follow_up_paper}

\end{document}